%% file: main.tex
\documentclass{article}
\usepackage{spconf,amsmath,graphicx}
\usepackage{booktabs,threeparttable}
\usepackage{standalone}
\usepackage{multirow}

\usepackage{amssymb}
\usepackage{color}
\usepackage[normalem]{ulem}
\title{adversarial training of end-to-end speech recognition\\ using a criticizing Language Model}

\name{Alexander H. Liu\qquad Hung-yi Lee\qquad Lin-shan Lee}
\address{College of Electrical Engineering and Computer Science\\
         National Taiwan University\\
         \small{\texttt{\{r07922013, hungyilee, lslee\}@ntu.edu.tw}
         }}

\begin{document}
%
\maketitle
\begin{abstract}
In this paper we proposed a novel Adversarial Training (AT) approach for end-to-end speech recognition using a Criticizing Language Model (CLM).
In this way the CLM and the automatic speech recognition (ASR) model can challenge and learn from each other iteratively to improve the performance.
Since the CLM only takes the text as input, huge quantities of unpaired text data can be utilized in this approach within end-to-end training.
Moreover, AT can be applied to any end-to-end ASR model using any deep-learning-based language modeling frameworks, and compatible with any existing end-to-end decoding method.
Initial results with an example experimental setup demonstrated the proposed approach is able to gain consistent improvements efficiently from auxiliary 
text data under different scenarios.
\end{abstract}

\input{intro.tex}
\input{method/method.tex}

\input{experiment/exp.tex}

\input{conclusion.tex}

\vfill\pagebreak

\bibliographystyle{IEEEbib}
{\small\bibliography{strings,refs}}

\end{document}

%% file: intro.tex
\section{Introduction}
\label{sec:intro}


With the fast advances of deep learning technologies, converting the well matured multi-module speech recognition processes~\cite{jelinek1976continuous} into a single speech-to-text model~\cite{graves2013speech} becomes highly attractive.
Such end-to-end speech recognition approaches are primarily based on two distinct models: connectionist temporal classification (CTC)~\cite{graves2006connectionist,graves2014towards,soltau2017neural} and sequence-to-sequence (Seq2seq)~\cite{bahdanau2016end,chorowski2015attention,chan2016listen} models.
By introducing an additional blank symbol and a specially defined loss function aggregating many allowed paths within a graph, CTC model can be optimized to generate the correct character sequences from the speech signals regardless of the blank symbols  interspersed among.
The seq2seq models, on the other hand, simply maximized the likelihood of observing the decoded sequence given the ground truth at every time step.
With many recent results~\cite{amodei2016deep,prabhavalkar2017comparison,chiu2017state,kim2017joint,hori2017advances} approaching the state-of-the-art, end-to-end deep learning has definitely been a very important direction for speech recognition.

Most end-to-end speech recognition approaches require a considerable amount of paired audio-text data, which is costly and time-consuming.
Semi-supervised approaches~\cite{thomas2013deep,vesely2013semi,dikici2016semi,karita2018semi} have been developed to address such problem by involving unpaired text data (which are relatively easy to obtain) in the training progress. 
One approach is to utilize unpaired text data to produce a separately trained language model (LM) to rescore the output of the end-to-end approach~\cite{mikolov2010recurrent,hori2017advances,chorowski2016towards,sriram2017cold}, but at the price of extra computation during testing.
Also, in this way the unpaired text data and paired audio-text data were used separately, and the machines could not learn from them jointly.
Another approach is to back-translate (synthesize) speech signals or encoder state sequences~\cite{karita2018semi,hayashi2018back,tjandra2017listening} from the unpaired text data, so they can be jointly learned in training. 
However, the improvements achievable with such approaches were limited by the quality of the synthesized data, which is usually far from real.
\input{figs/fig1.tex}

The Generative Adversarial Networks (GANs)~\cite{goodfellow2014generative} have been shown to be very successful in diversified application areas.
Instead of learning from a set of ground truth taken as the upper bound for learning, a generator model and a discriminator model are trained iteratively to challenge and learn from each other step by step.
In this paper, inspired by GANs, we propose a novel approach to embed the advantages of adversarial training (AT) into end-to-end speech recognition. With the proposed approach, huge quantities of unpaired text data can be utilized without a separately trained model, extra computation during testing and the shortcomings of back-translation style data augmentation. 

%% file: figs/fig1.tex
\begin{figure}[t]
\centerline{\includegraphics[width=8.7cm]{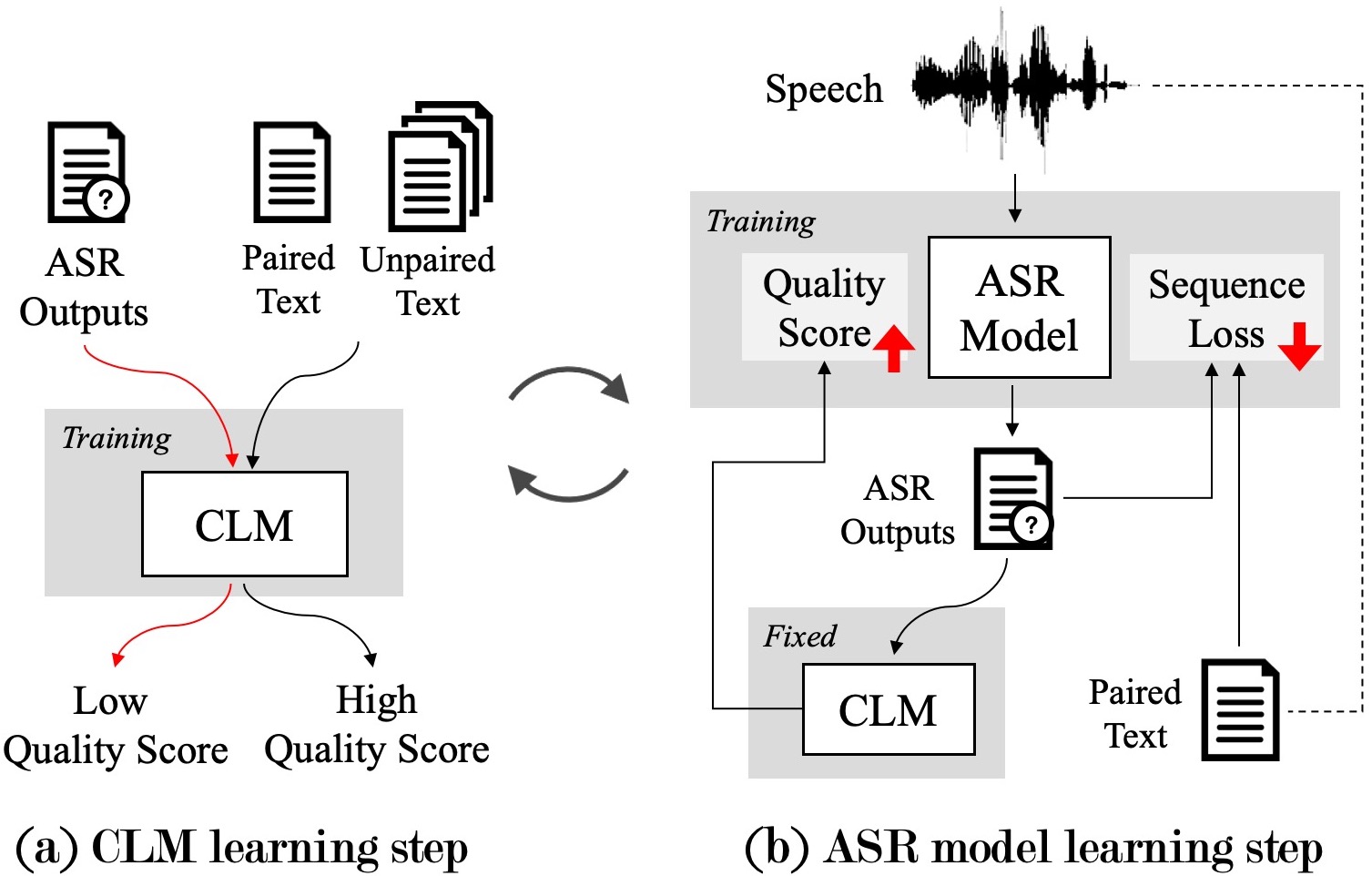}}
\vspace{-7pt}
\caption{\small Overview of the Adversarial Training (AT) approach for end-to-end speech recognition. The two steps here are conducted iteratively: (a) a Criticizing Language Model (CLM) is trained to evaluate the quality score given a text sequence, and (b) and ASR model is trained to minimize the sequence loss calculated with ground truth while maximizing the scores given by CLM.}

\label{fig:overview}
\end{figure}

%% file: method/method.tex
\section{Proposed Approach}
\label{sec:method}

\input{method/overview.tex}

\input{method/alm.tex}
\input{method/asr.tex}

%% file: method/overview.tex
\vspace{-5pt}
\subsection{Overview}
\vspace{-3pt}
\label{subsec:overview}
In our Adversarial Training approach to end-to-end speech recognition, the ASR model is considered as a generator conditioned on the input speech signal whose output is the corresponding transcription.
A \textit{Criticizing Language Model} (CLM) is used as a discriminator to distinguish real text from ASR transcriptions.
The ASR model and CLM are trained iteratively, so they learn from each other step by step.
Fig.~\ref{fig:overview} gives an overview of the proposed approach.

In Fig.~\ref{fig:overview}(a) for CLM training step, the CLM learns to assign higher scores to real text and lower scores to ASR transcriptions.
The real text here does not have to be paired with audio, which is how the unpaired text can be involved in the training processes.
This CLM is to evaluate the quality of each given text sequence by offering a score for adversarial purposes, with details given in Sec.~\ref{subsec:alm}.

In Fig.~\ref{fig:overview}(b) for ASR model learning step, the parameters of CLM are fixed and we train the ASR model by minimizing the sequence loss (e.g. seq2seq loss and/or CTC loss) evaluated with the ground truth just as typical end-to-end training.
At the same time, with CLM acting as a discriminator evaluating the quality score for the output of ASR model, the ASR model also has to learn to generate transcriptions obtaining higher quality scores from CLM.
The details of the ASR model is in Sec.~\ref{subsec:asr}. 

Note that the ASR model and the CLM are learned iteratively both  from scratch. No pre-training is needed.
Each of them improves itself based on the challenges offered by the other in each iteration.
Once the training ends, the ASR model is expected to implicitly leverage the linguistic knowledge learned from CLM, and the latter is no longer used during testing. 
This approach can be used with any existing end-to-end speech recognition frameworks and any language modeling framework.
Below we take one example set of the proposed approach to explain the details.

%% file: method/alm.tex
\vspace{-9pt}
\subsection{Criticizing Language Model (CLM)}
\label{subsec:alm}
\input{figs/fig2.tex}

\noindent\textbf{Network Architecture.}
CLM takes either real text or ASR transcriptions as input and outputs a scalar $s$ as the quality score.
The real text is represented as a sequence of one-hot vectors $y=y_1,y_2,...,y_L$, while for ASR transcriptions this is a sequence of vectors for distributions $\tilde{y}=\tilde{y_1},\tilde{y_2},\tilde{y_3},...$ .
Fig.~\ref{fig:alm} illustrates an example architecture of CLM used in this work. 
The input vector sequence $y$ (or $\tilde{y}$) is first projected to a lower dimensional space through a single layer neural net. 
Next, two layers of one-dimensional convolution neural network extracts features for each time index. 
Finally, average pooling over the time axis is applied to get a single representative feature, which is then transformed to a scalar $s$ (the quality score) with linear projection.

The reason a convolution-based network instead of a recurrent network is used in Fig~\ref{fig:alm} is twofold. 
Convolution with small window size captures local relation features, which can then be averaged over time.
Also, CNN based network is relatively more computationally efficient, which is important in adversarial training. 
But other network architectures such as RNN-LM~\cite{hori2017advances} can also be used here.

\noindent\textbf{Loss Function.}
A major problem here is that soft distribution vectors produced by the ASR model is very different from one-hot vectors for real text data, making the task of CLM trivial, and the ASR model almost always fail to compete against it.
Thanks to Wasserstein GAN (WGAN)~\cite{arjovsky2017wasserstein} which addressed the above problem to some good extent.
Based on the concept of WGAN, CLM is designed to estimate the \textit{Earth-Mover} (Wasserstein-1)~\cite{villani2008optimal} distance between sequences from real data and ASR output. 
The loss function of CLM is the weighted sum of a loss $L_{D}$ and a gradient penalty $gp$ as follows, 
\vspace{-5pt}
\begin{equation}
L_{CLM} =  \lambda_{CLM} L_{D} + \lambda_{gp}~gp,
\label{eq:L_ALM}
\vspace{-3pt}
\end{equation}
in which $\lambda_{CLM}$,$\lambda_{gp}$ are wieghts and $L_{D}$ and $gp$ are respectively in Eq~(\ref{eq:dist}) and Eq~(\ref{eq:gp}) below. 
\begin{equation}
\label{eq:dist}
L_{D} = \mathop{\mathbb{E}}_{\tilde{y}\thicksim \mathbb{P}_{a}}\big[CLM(\tilde{y})\big] -
\mathop{\mathbb{E}}_{y\thicksim \mathbb{P}_{d}}\big[CLM(y)\big],
\vspace{-3pt}
\end{equation}
where $CLM(y)$ is the quality score for $y$ given by CLM,
$\mathbb{P}_{a}$ the distribution of ASR output $\tilde{y}$ and
$\mathbb{P}_{d}$ the distribution of real text $y$. 
The 1-Lipschtiz restriction is imposed for CLM by applying the gradient penalty~\cite{gulrajani2017improved} as below, 
\begin{equation}
\label{eq:gp}
gp = \mathop{\mathbb{E}}_{\hat{y}\thicksim\mathbb{P}_{\hat{y}}} {\big[(\lVert\nabla_{\hat{y}}CLM(\hat{y})\rVert-1)^2\big]},
\vspace{-3pt}
\end{equation}
where $\hat{y}$ are samples generated by randomly interpolating between $\tilde{y}$ and $y$,
and $\mathbb{P}_{\hat{y}}$ is the distribution of $\hat{y}$.

%% file: figs/fig2.tex
\begin{figure}[h]

\centerline{\includegraphics[height=6cm]{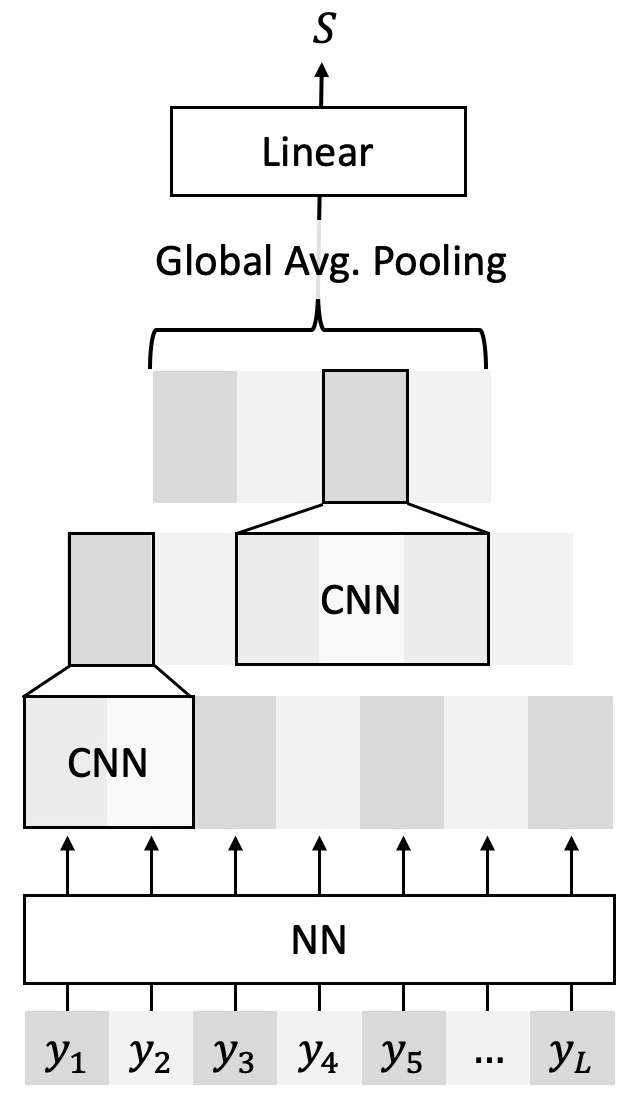}}
\vspace{-9pt}
\caption{Network architecture of the CLM.}
\label{fig:alm}
\end{figure}

%% file: method/asr.tex
\vspace{-3pt}
\subsection{ASR Model}
\label{subsec:asr}
\input{figs/fig3.tex}

\textbf{Network Architecture.}
Any network architecture for end-to-end speech recognition can be used here, while Fig.~\ref{fig:asr} gives the one used in this work, following the previous work~\cite{hori2017advances} of integrating attentioned Seq2seq with CTC.
The model takes a sequence of speech features $O=o_1,o_2,...,o_N$ with length $N$ as the input.
$O$ is encoded into sequence of hidden state $H=h_1,h_2,...,h_{T}$ by the encoder (consists of a VGG extractor performing input downsampling followed by several BLSTM layers) with $T$ being the output sequence length.
The decoder is a single layer LSTM maintaining its own hidden state $q$.
For each time index $t$, location-aware attention mechanism~\cite{chorowski2015attention} $Attention$ is used to integrate $H$ with the previous decoder state $q_{t-1}$ to generate the context vector $c_t$.
The decoder then decodes $c_t$ together with the ground truth one-hot vector of the previous time step $y_{t-1}$ to $q_t$.
Finally, a fully connected layer with softmax activation $CharDistribution$ takes $q_t$ and predicts the distribution vector $\tilde{y_t}$.
$h_t$ is also projected to $\check{y_t}$ with linear layer $Transform$ as output to help in learning of the encoder as shown in previous work~\cite{kim2017joint}.
The ASR model outputs two character sequences, $\tilde{y}=\tilde{y_1},\tilde{y_2},...,\tilde{y_T}$ and $\check{y}=\check{y_1},\check{y_2},...,\check{y_T}$, respectively supervised by Seq2seq loss and CTC loss.
CLM only takes $\tilde{y}$ as input. 
During testing, $\tilde{y}$ and $\check{y}$ are integrated into a single output sequence just as done in the previous work~\cite{hori2017advances}.
\noindent\textbf{Seq2seq Loss.}
The Seq2seq ASR model is to estimate the posterior probability,
\vspace{-5pt}
\begin{equation}
\label{eq:Seq2seq}
P_{s2s}(\tilde{y}|O) = \prod_{l=1}^{T}{P_{s2s}(\tilde{y}_l|\tilde{y}_{1:l-1},O)}.
\end{equation}

The loss function of the Seq2seq model model can be then computed as below,
\begin{equation}
\label{eq:loss_att}
L_{s2s} \equiv -\log P_{s2s}(y|O) = - \sum\limits_{t=1}^{T} \log P_{s2s}(y_t|y_{1:t-1},O),
\end{equation}
except here $y=y_1,y_2,...,y_T$ is the ground truth of $O$ with length $T$.

\textbf{CTC Loss.}
CTC~\cite{graves2006connectionist} objective function is also used in this work for multi-task learning. 
CTC computes the posterior probability as below, 
\begin{equation}
P(y|O) = - \sum\limits_{\pi \in y'} P(\pi|O),
\vspace{-6pt}
\end{equation}
where $y'$ is the set of all possible sequences $\pi$ obtained by arbitrarily repeating symbols of $y$ and inserting blank symbols into $y$.
The probability $P(\pi|O)$ can be approximated by $\check{y}$,
\vspace{-5pt}
\begin{equation}
P(\pi|O) \approx  \prod_{t=1}^{T}{P_{ctc}(\check{y}_t|O)}.
\vspace{-6pt}
\end{equation}
The loss function of CTC is defined as:
\begin{equation}
\label{eq:loss_ctc}
L_{ctc} \equiv -\log P(y|O).
\vspace{-3pt}
\end{equation}

\noindent\textbf{Total Loss.}
The ASR model is trained by minimizing the loss function constructed with Eq~(\ref{eq:loss_att}) and (\ref{eq:loss_ctc}) minus the quality score from CLM,
\begin{equation}
\label{eq:asr_loss}
L_{ASR} = \lambda_{s2s}~L_{s2s} + (1-\lambda_{s2s})~L_{ctc} - \lambda_{CLM}~CLM(\tilde{y})
\end{equation}
where  $\lambda_{s2s}$ controls the weights for the multi-task learning between Seq2seq and CTC. 
The last term in Eq~(\ref{eq:asr_loss}) is the adversarial loss from CLM, pushing the ASR model to maximize the quality score from CLM.

%% file: figs/fig3.tex
\begin{figure}[h]

\centerline{\includegraphics[height=6.5cm]{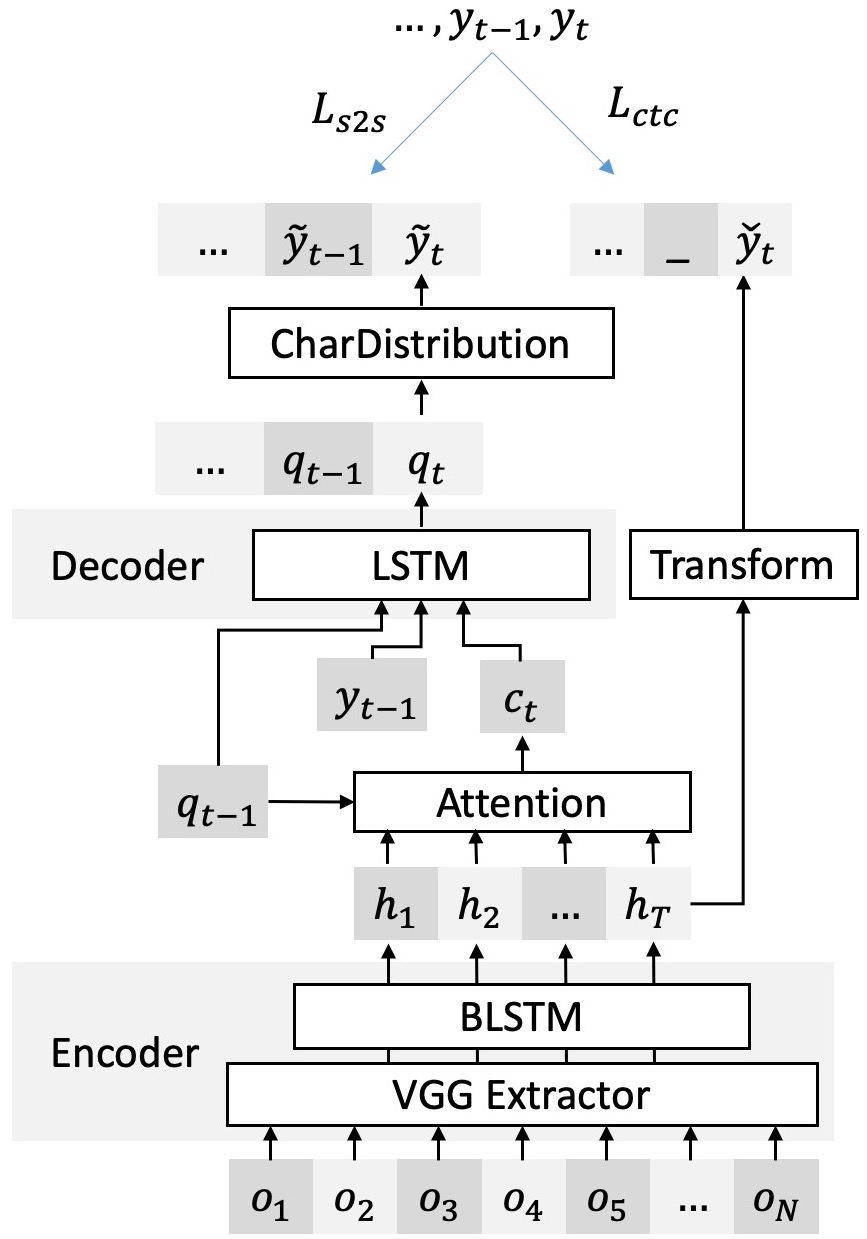}}
\vspace{-4pt}
\caption{Network architecture of the ASR model.}
\label{fig:asr}
\end{figure}

%% file: experiment/exp.tex
\vspace{-8pt}
\section{Experiment}
\label{sec:experiment}
\input{experiment/setting.tex}
\input{tables/performance.tex}
\input{experiment/qn_result.tex}

\input{experiment/ql_result.tex}

\input{tables/demo.tex}

%% file: experiment/setting.tex
\vspace{-3pt}
\subsection{Experimental Setup}
\label{subsec:setting}
\vspace{-3pt}
The experiments were performed on the  LibriSpeech~\cite{panayotov2015librispeech}.
100 hours of clean speech data and their transcriptions are used as the paired data. 
We took the text of other 360 hours of clean speech and 500 hours of noisy speech and utilized them as the unpaired data (text-only). 
The clean development set and clean test set were used for evaluation. 
We used the end-to-end speech processing toolkit ESPnet~\cite{watanabe2018espnet} for data preprocessing  and customized it for our adversarial training processes.
We followed the previous work~\cite{hori2017advances,hayashi2018back} to use 80-dimensional log Mel-filter bank and 3-dimensional pitch features as the acoustic features.
Text data are represented by sequences of 5000 subword units one-hot vectors.
For the CLM model, the dimension of the output of all layers were set to 128 except the last.
The first convolution had a window size of 2 and stride of 1, and the second had window size 3 and stride 1. 
Batch normalization is applied between layers.
For the ASR model, the encoder included a 6-layer VGG extractor with downsampling used in the previous work~\cite{hori2017advances} and a 5-layer BLSTM with 320 units per direction. 300-dimensional location-aware attention~\cite{chorowski2015attention} was used in the attention layer.
The decoder was a single layer LSTM with 320 units.
$\lambda_{gp}$ was set to 10 and $\lambda_{s2s}$ is set to 0.5.
$\lambda_{CLM}$ is set to $10^{-4}$ since CLM output value was usually much higher than other loss values.
Also, the update frequency of CLM is set to 5 times less than the ASR model to stabilize AT process.

%% file: tables/performance.tex
\begin{table}[t]
\small
\centering
\begin{threeparttable}

\caption{\small Speech recognition performance. "+LM" refers to shallow fusion decoding jointly with RNN-LM~\cite{hori2017advances}, "+AT" refers to the adversarial training proposed here, "+Both" indicates training with AT and joint decoding with RNN-LM, and BT is the prior work of back-translation~\cite{hayashi2018back}.}

\begin{tabular}{c| l|c c | c}
\toprule
\multirow[c]{2}{*}{Data}    &     \multirow[c]{2}{*}{~~Method}     &   \multicolumn{2}{c|}{CER/WER (\%)}  & WER $\Delta ^\dag$\\ 
     &          & Dev           & Test & Test  \\ \hline 
\multirow[c]{4}{*}{\shortstack{(A)\\w/o\\unpair\\text}} & (a) { Baseline}                   & { 10.5 / 21.6}   & 10.5 / 21.7    & -     \\ 
& (b)~{\small +LM}                           & 10.9 / 20.0   & 11.1 / 20.3       &  ~~6.5\%    \\ 
& (c)~{\small +AT}                              & \textbf{~9.5 / 19.9}	&  \textbf{~9.6 / 20.1}    &  \textbf{~~7.4\%}   \\ 
& (d)~{\small +Both}                       &  ~9.4 / 17.9   &  ~9.7 / 18.3          &  15.7\%  \\ \hline
\multirow[c]{5}{*}{\shortstack{(B)\\  w/\\360hrs\\text}}
& (e) {\small +LM}                           & 10.5 / 19.6   & 10.6 / 19.6       &   ~~9.7\%      \\ 
& (f) {\small +AT}                              &  \textbf{~9.1 / 19.1}   &  \textbf{~9.5 / 19.2}    &   \textbf{11.5\%}      \\
& (g) {\small +Both}                       &  ~9.0 / 17.1   &  ~9.1 / 17.3          &   20.3\%      \\ 
& (h) ${\text{BT}}^{\ddag}$    &  10.3 / 23.5    & 10.3 / 23.6       &    ~~6.3\%            \\
& (i)  $\text{BT+LM}^{\ddag}$                      & ~9.8 / 21.6 & 10.0 / 22.0           &    12.7\%  \\\hline
\multirow[c]{3}{*}{\shortstack{(C)~w/\\860hrs\\text}}
& (j)~+LM       & ~9.9 / 18.6   &  10.2 / 18.8     &   13.4\%    \\ 
& (k)~+AT       &  \textbf{~8.6 / 18.5}   &   \textbf{~8.8 / 18.7}     &   \textbf{13.8\%}    \\
& (l)~+Both     & ~7.9 / 15.3   &   ~8.2 / 15.8     &   27.2\%    \\ 
\bottomrule
\end{tabular}
\vspace{-3pt}
\begin{tablenotes}
\item{\text{\dag}} \text{~{\small Relative improvement with respect to the baseline.}}
\item{\text{\ddag}} \text{~{\small Prior work~\cite{hayashi2018back}, baseline WER 25.2\% on test set reported.}}
\end{tablenotes}
\label{table:comp}
\end{threeparttable}
\vspace{-3pt}
\end{table}

%% file: experiment/qn_result.tex
\subsection{Experimental Results}
\label{subsec:qn_result}
\vspace{-5pt}
In the experiments, the ASR model was trained on the 100 hours speech data but combined with different amount of unpaired text utilized in different ways. 
The results are listed in Table~\ref{table:comp}, where "Baseline" refers to the plain end-to-end speech recognition framework as described in Sec.~\ref{subsec:asr},
"+LM" refers to the shallow fusion decoding with a separately trained RNN language model (RNN-LM)~\cite{hori2017advances,sriram2017cold} and "+AT" refers to the adversarial training proposed here.
AT is actually compatible with any existing end-to-end speech recognition decoding approach, so "+Both" refers to training with AT while jointly decoding with RNN-LM.
We ran all experiments three times with random initialization and reported the averaged error rate with decoding beam size set to 20.

Part (A) lists the results without extra text data.
It is  worth mentioning that even without extra text data, AT offered improvements over the baseline (rows(c) vs (a)), and the performance was further improved when integrated with RNN-LM (rows(d) vs (c)).  
Parts (B) and (C) are for results respectively with 360 hours and 860 hours of unpaired text data.
We see AT lowers recognition error rate as the RNN language model do (rows(f) vs (e), (k) vs (j)) and the improvements can be accumulated (rows(g) vs (f), (l) vs (k)).
The previous work of back-translation (BT) style data augmentation~\cite{hayashi2018back}, which aimed to utilize unpaired text data as AT do, was also listed in rows (h),(i).
We see AT did better than BT under the same setting (rows (f) vs (h) and (g) vs (i)).

Fig~\ref{fig:beam} demonstrates the performance gap between different models (rows (a), (e), (f) and (g) of Table~\ref{table:comp}) with varying the beam size from 1 to 30.
The points for beam size 20 are those in Table~\ref{table:comp}.
We see that the proposed AT consistently improved the performance regardless of the beam size during decoding.
It is clear that for all beam sizes considered AT outperformed RNN-LM in terms of utilizing the extra text data (curves (f) vs (e)), and AT is compatible to  and able to offer additional improvements on top of the separately trained RNN-LM (curves (g) vs (f)).
All these verified AT proposed here is able to integrate more linguistic knowledge from unpaired text data into the ASR model.

%% file: experiment/ql_result.tex
\input{figs/fig4.tex}
Table~\ref{table:demo} provides some transcriptions obtained with the four models shown in rows (a)(e)(f)(g) of Table~\ref{table:comp} on the same input utterances from the testing set. 
All models were trained with 100 hours of paired data, while the lower three with additional text of 360 hours, all with beam size 20. 
We see that AT seemed to make the output more grammatical.
In the first example, AT is able to predict the correct words. 
In the second example, although all the models misrecognized the word ``Alexander'', the transcriptions provided by models with AT (rows (f)(g)) are more grammatical.

%% file: figs/fig4.tex
\begin{figure}[t]
\centerline{\includegraphics[height=4.7cm]{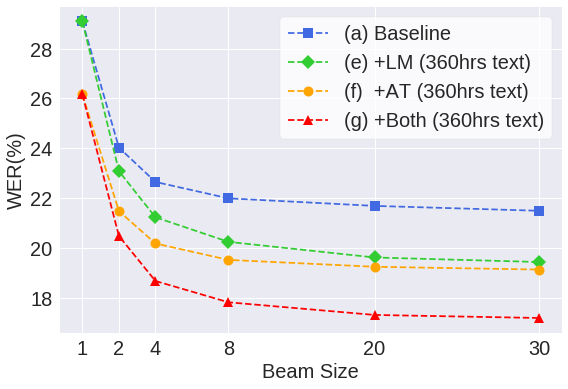}}
\vspace{-10pt}
\caption{\small Testing set WER for varying beam size. Index of curve  shared with the corresponding row in Table 1.}

\label{fig:beam}
\end{figure}

%% file: tables/demo.tex
\begin{table}[htb]
\vspace{-11pt}
\small
\centering
\caption{\small Transcription examples, with ASR errors in uppercase and the differences made by AT underlined. Index shared with Table~\ref{table:comp}.}
\begin{tabular}{l| l}
\toprule
Model & Transcription\\\hline 
Truth       & nonsense of course i can't really ...\\
(a) Baseline    & NON SENSE of course i CAN'TVERLY  ...\\
(e) +LM      & NON SENSE of course i can't FREELY ...\\
(f) +AT        & \underline{nonsense} of course i \underline{\smash{can't really}} ...\\
(g) +Both       & \underline{nonsense} of course i \underline{\smash{can't really}} ...\\
\hline 
Truth       & alexander did not sit down \\
(a) Baseline    & OUTSIDEED IT not SET down	\\
(e) +LM      & WHY did not sit down\\
(f) +AT        & \underline{ALICE} did not sit down\\
(g) +Both       & \underline{ALICE} did not sit down\\

\bottomrule
\end{tabular}
\label{table:demo}
\vspace{-10pt}
\end{table}

%% file: conclusion.tex
\vspace{-5pt}
\section{Conclusion}
\label{sec:conclusion}
\vspace{-5pt}
In this paper we proposed a novel framework for adversarial training end-to-end speech recognition using a criticizing language model.
This offers a direction for better utilizing additional text data without the need for a separately trained language model.
This framework can be used with any end-to-end speech recognition and language modeling frameworks.
Experiments on one example set of the proposed framework showed consistent improvement over different settings.